%% file: 0-main.tex
\documentclass{article}
\usepackage{ijcai22}

\usepackage{times}
\usepackage{soul}
\usepackage{url}
\usepackage[hidelinks]{hyperref}
\usepackage[utf8]{inputenc}
\usepackage[small]{caption}
\usepackage{graphicx}
\usepackage{amsmath}
\usepackage{amsthm}
\usepackage{booktabs}
\usepackage{algorithm}
\usepackage{algorithmic}
\title{Towards Label-Efficient Incremental Learning: A Survey}

\author{
Mert Kilickaya$^1$,
Joost van de Weijer$^2$\And
Yuki M. Asano$^3$
\affiliations
$^1$Eindhoven University of Technology\\ 
$^2$Autonomous University of Barcelona\\
$^3$University of Amsterdam
\emails
m.kilickaya@tue.nl,
joost@cvc.uab.es,
y.m.asano@uva.nl
}

\input{style}

\begin{document}

\maketitle

\begin{abstract}

The current dominant paradigm when building a machine learning model is to iterate over a dataset over and over until convergence. Such an approach is non-incremental, as it assumes access to all images of all categories at once. However, for many applications, non-incremental learning is unrealistic. To that end, researchers study incremental learning, where a learner is required to adapt to an incoming stream of data with a varying distribution while preventing forgetting of past knowledge. Significant progress has been made, however, the vast majority of works focus on the fully supervised setting, making these algorithms label-hungry thus limiting their real-life deployment. To that end, in this paper, we make the first attempt to survey recently growing interest in label-efficient incremental learning. We identify three subdivisions, namely semi-, few-shot- and self-supervised learning to reduce labeling efforts. Finally, we identify novel directions that can further enhance label-efficiency and improve incremental learning scalability. Project website: \href{https://github.com/kilickaya/label-efficient-il}{https://github.com/kilickaya/label-efficient-il}.

\end{abstract}

\input{1-intro.tex}


\input{2-background.tex}
\input{3-semi.tex}

\input{4-few.tex}
\input{5-self.tex}
\input{6-conclusion.tex}

\bibliographystyle{named}
\bibliography{ijcai22}

\end{document}

%% file: style.tex
\usepackage{times}
\usepackage{epsfig}
\usepackage{graphicx}
\usepackage{amsmath}
\usepackage{amssymb}

\usepackage{bbm}

\usepackage{amssymb}
\usepackage{slashbox}
\usepackage[dvipsnames,table]{xcolor} 
\usepackage{tabularx,ragged2e}
\usepackage{spreadtab}
\usepackage{adjustbox}
\usepackage{booktabs}
\usepackage{xcolor}
\usepackage{multirow}
\usepackage{rotating}
\usepackage{color, colortbl}

\usepackage[inline]{enumitem}
\usepackage{color,soul}
\usepackage[caption=false]{subfig}

\graphicspath{{./figures/}}

\DeclareGraphicsExtensions{.jpeg,.png,.eps,.pdf}

\newcolumntype{a}{>{\columncolor{Gray}}c}

\newcommand{\partitle}[1]{\bigbreak\noindent\textbf{#1}}

\makeatletter
\newcommand*{\rom}[1]{\expandafter\@slowromancap\romannumeral #1@}
\makeatother

\usepackage{mathtools}

\usepackage{pifont}
\newcommand{\xmark}{\ding{55}}%

\newlist{todolist}{itemize}{2}
\setlist[todolist]{label=$\square$}
\usepackage{pifont}

\makeatletter
\newcommand*\bigcdot{\mathpalette\bigcdot@{.5}}
\newcommand*\bigcdot@[2]{\mathbin{\vcenter{\hbox{\scalebox{#2}{$\m@th#1\bullet$}}}}}
\makeatother

\definecolor{applegreen}{rgb}{0.55,0.71,0.0}
\definecolor{babyblue}{rgb}{0.54,0.81,0.94}
\definecolor{azure}{rgb}{0.0,0.5,1.0}
\definecolor{budgreen}{rgb}{0.48,0.71,0.38}
\definecolor{amaranthpurple}{rgb}{0.67,0.15,0.31}

\newcommand{\ie}{\textit{i}.\textit{e}., }
\newcommand{\eg}{\textit{e}.\textit{g}. }

\usepackage{cleveref}

\crefformat{section}{\S#2#1#3} 
\crefformat{subsection}{\S#2#1#3}
\crefformat{subsubsection}{\S#2#1#3}

\usepackage{booktabs,arydshln}

\makeatletter
\def\adl@drawiv#1#2#3{%
        \hskip.5\tabcolsep
        \xleaders#3{#2.5\@tempdimb #1{1}#2.5\@tempdimb}%
                #2\z@ plus1fil minus1fil\relax
        \hskip.5\tabcolsep}
\newcommand{\cdashlinelr}[1]{%
  \noalign{\vskip\aboverulesep
           \global\let\@dashdrawstore\adl@draw
           \global\let\adl@draw\adl@drawiv}
  \cdashline{#1}
  \noalign{\global\let\adl@draw\@dashdrawstore
           \vskip\belowrulesep}}
\makeatother

\definecolor{lightcyan}{rgb}{0.88,1,1}
\definecolor{applegreen}{rgb}{0.55, 0.71, 0.0}
\definecolor{aqua}{rgb}{0.0, 1.0, 1.0}
\definecolor{beaublue}{rgb}{0.74, 0.83, 0.9}
\definecolor{blond}{rgb}{0.98, 0.94, 0.75}
\definecolor{caribbeangreen}{rgb}{0.0, 0.8, 0.6}
\definecolor{classicrose}{rgb}{0.98, 0.8, 0.91}
\definecolor{darkseagreen}{rgb}{0.56, 0.74, 0.56}
\definecolor{lightgreen}{rgb}{0.56, 0.93, 0.56}
\definecolor{mediumaquamarine}{rgb}{0.4, 0.8, 0.67}
\definecolor{babypink}{rgb}{0.96, 0.76, 0.76}
\definecolor{cambridgeblue}{rgb}{0.64, 0.76, 0.68}
\definecolor{celadon}{rgb}{0.67, 0.88, 0.69}
\definecolor{etonblue}{rgb}{0.80, 0.92, 0.64}

%% file: 1-intro.tex
\section{Introduction}



Deep learning is the dominant approach to build highly performant machine learning systems that are deployed in a wide range of scenarios – from  self-driving cars to mobile applications. 
To this end, the classic approach is to choose a current state-of-the-art neural network architecture, such as a ResNet~\cite{he2016deep} or a Vision-Transformer~\cite{dosovitskiy2020image} and combine this with large-scale datasets containing millions of annotations, such as ImageNet. 
The learning of the model is done supervisedly by iterating over the dataset multiple times until some convergence or stopping criterion is met. 
We name such approaches \textit{non-incremental}, as the learner assumes access to the whole data at all times. 

\textit{Non-incremental learning}, which implicitly assumes a static world, has severe limitations with regard to its performance and applicability: First, the list of categories the model can distinguish from is fixed. 
Second, once the model is deployed, the model no longer makes use of the ever-growing data with or without labels to self-improve. 
Third, the ability to iterate over all the data may be impossible due to privacy regulations or data storage legislation.  



\begin{figure*}[t]
    \centering
\includegraphics[width=0.88\textwidth]{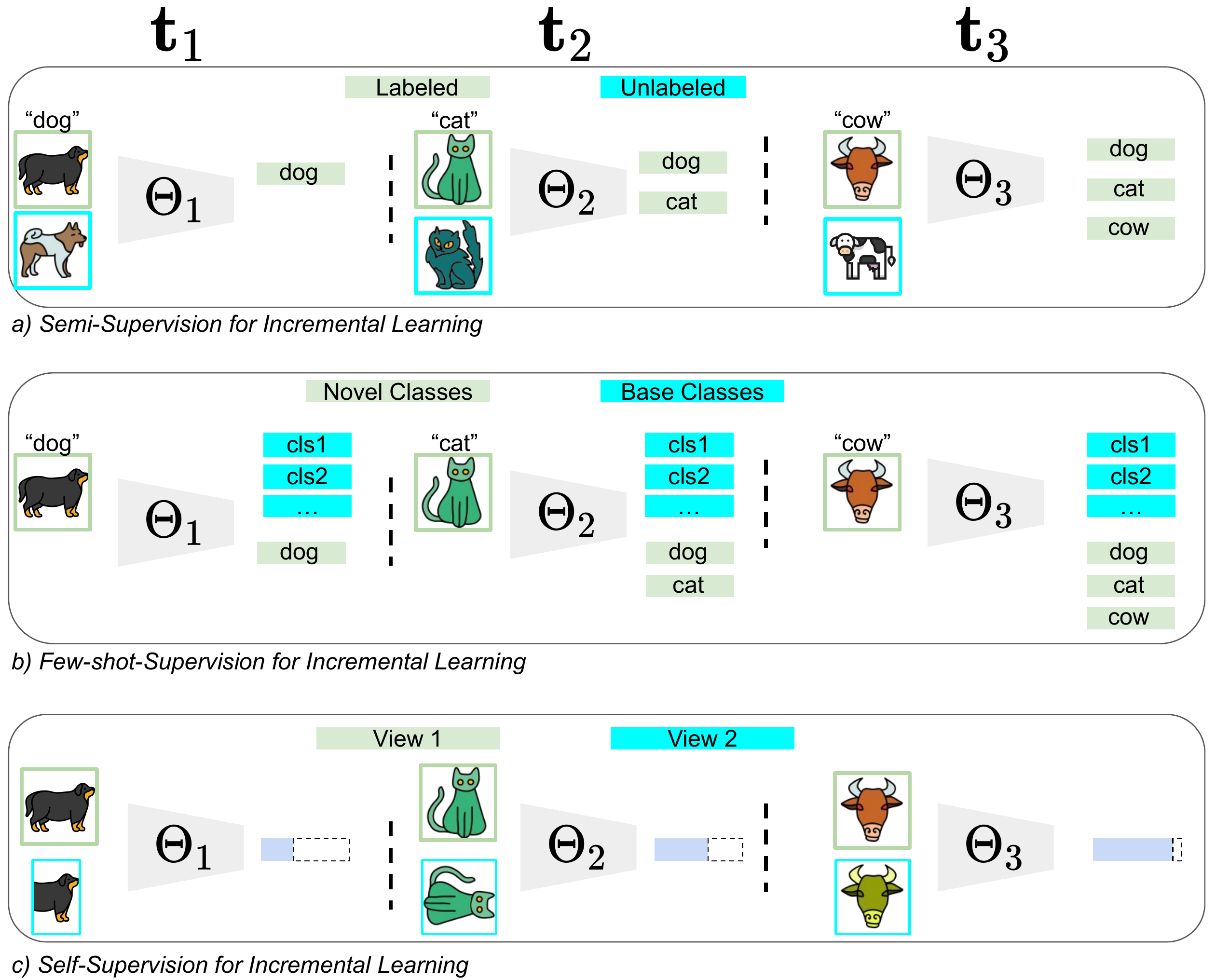}


 \caption{In incremental learning, a disjoint set of classes (a task) arrive sequentially. Then, the learner is optimized to distinguish across these categories. Here, we provide three different forms of label-efficient incremental learning. \textit{a).} Semi-Supervision pseudo labels unlabeled data to reduce the need for labeled data, \textit{b).} Few-shot-Supervision expands a pre-trained classifier with only few labeled instances (\ie 1-shot), \textit{c).} Self-Supervision designates pre-text tasks to completely remove the need for human labels to train incremental learners. Here, the goal is to accumulate representations by matching different views (augmentations such as cropping, rotation and color jitter) of the same input image through incremental learning. Best viewed in color.}
    
    \label{fig:teaser}
\end{figure*}

Motivated by these challenges, researchers put increasing attention on incremental (or continual) learning~\cite{masana2020class}. In incremental learning, the learner receives the learning tasks sequentially (\eg first dog, then cat, and then finally cow, see Figure~\ref{fig:teaser}), whose data then disappears after some iterations. This way, a deep classifier can be updated with novel data, while preserving the performance on the previous tasks. Yet, despite its clear setting and large potential for scalable applications, why is the incremental learning paradigm not used as often as the non-incremental counterpart? 

One reason is often cited is the phenomenon of catastrophic forgetting~\cite{french1999catastrophic}. The performance of the model on the previous tasks deteriorates while observing more incremental tasks. However, thanks to efforts in regularizing neural network weights from abruptly drifting~\cite{li2017learning,zenke2017continual}, or replaying previous data from the memory~\cite{shin2017continual}, forgetfulness has reduced dramatically in years. 
In this paper, we raise the attention of the incremental learning researchers to an equally important, yet up to now unsolved, issue, \textit{Scalability}. In particular, incremental learners are extremely annotation-hungry, as they demand large amounts of labeled data for achieving comparable performances. 



The importance of \textit{label-efficient} learning generally is well recognized in non-incremental learning, as researchers organize dedicated workshops on the topic, see~\cite{L2ID}. Success has been made since researchers can build upon large-scale pre-trained models to transfer learning on their individual tasks with limited data~\cite{radford2021learning}. 

However, the importance of label-efficiency specifically for incremental learning is yet to be recognized and addressed. While there have been few studies that surveys conventional incremental learners~\cite{masana2020class,de2021continual,wang2023comp,zhou2023class}, their focus is still on the old label-hungry paradigm in incremental learning.  This motivates our effort in surveying recent techniques to build large-scale, label-efficient incremental learners. 

We identify three main directions, see Figure~\ref{fig:teaser}. Semi-supervised learners combine limited labeled data with abundant unlabeled data to reduce label-supervision via pseudo supervision~\cite{wang2021ordisco}. Few-shot-supervised learners reduce label-supervision by expanding a pre-trained classifier with only few examples (\eg 1-shot)~\cite{tao2020few}. Finally, self-supervised learners omit the need for label-supervision by designing pretext tasks from the incremental learning tasks~\cite{fini2022self}. 

Our survey first introduces the necessary notation in Section~\cref{sec:background}, and then focuses on Semi-Supervision (Section~\cref{sec:semi}), Few-shot-Supervision (Section~\cref{sec:few}) and Self-Supervision (Section~\cref{sec:self}). We then conclude with future research directions and novel problems in Section~\cref{sec:conclusion}. 




%% file: 2-background.tex
\input{table1.tex}

\section{Label-Efficient Incremental Learning}
\label{sec:background}



There are three main settings, namely task-, domain- and class-incremental learning, see~\cite{van2019three}. This short survey focuses on class-incremental learning as it received more advances in the literature, however the proposed subdivision can also be applied to other settings. 






\subsection{Incremental Learning} 


The main goal of a class-incremental learner is two-fold: \textit{i)} To learn to recognize novel classes in the current task, \textit{ii)} To preserve the performance on previously learned classes (\ie reduce forgetfulness). A class-incremental learning task is defined by the length of the learning sequence and the number of classes within each task (\ie Figure~\ref{fig:teaser} depicts a 3-step 1-class incremental learning setting). 


Formally, following the notation in~\cite{madaan2021representational}, the learner receives a sequence of learning tasks $\mathcal{T}_{1:t} = (\mathcal{T}_{1}, \mathcal{T}_{2}, ...,\mathcal{T}_{t})$, with a corresponding dataset $\mathcal{D}_{\mathcal{T}} = \{ (x_{i,t}, y_{i,t})_{i=1}^{n_{t}} \}$ with $n_{t}$ instances. Here, an input pair $\{x_{i,t}, y_{i,t}\} \in \mathcal{X}_{t} \times \mathcal{Y}_{t}$ is sampled from an unknown distribution. 



When the learning task arrives, then a deep convolutional feature extractor is optimized $f_{\Theta}: \mathcal{X}_{t}\rightarrow \mathcal{R}^{D}$ and a linear classifier $h_{\phi}: \mathcal{R}^{D}\rightarrow\mathcal{Y}_{t}$ jointly parameterized by 
$W=\{\Theta, \phi\}$. The feature extractor maps the input to a $D$-dimensional embedding space, and the linear classifier further projects the input to the class space, by typically optimizing the following objective: $\mathbf{CE}(h_{\phi}(f_{\Theta}(x_{i,t})), y_{i,t})$
\noindent where $\mathbf{CE}(\cdot)$ is the standard Cross-Entropy loss to classify the input. This way, the model is fine-tuned sequentially with the incoming stream of task data. 

\partitle{Mitigating Forgetting.} Fine-tuning with exemplars may lead to catastrophic forgetting of past tasks, since the only objective is classification. To that end, a simple technique is to only update a few layers closer to classifier head. Two fundamental ways to prevent forgetfulness are regularization and replay. Regularization often constraints neural network weights from abrupt drifts across learning tasks (\ie via penalizing the norm of change, simply by $ || W_{t-1} - W_{t}||$, see~\cite{zenke2017continual}). Replay often stores a subset of the learning task data within the memory to replay during incremental learning tasks, such as experience replay~\cite{rolnick2019experience}.

\partitle{Evaluation.} Incremental learners are often evaluated by accuracy (higher is better) and forgetfulness (lower is better).  \textit{i) Accuracy} measures the test accuracy for all the learned tasks until the task $t$ as $A_{t} = \dfrac{1}{t}\sum_{i=1}^{t} a_{t, i}$.  \textit{ii) Forgetfulness:} is the average performance decrease while learning incremental tasks, after the completion of the learning, as: $F = \dfrac{1}{T-1} \sum_{i=1}^{T-1} \max_{t\in \{1, ..., t\}}(a_{t,i} - a_{T, i})$ for $T$ total learning tasks. 




\subsection{Label-Efficient Learning}



\partitle{Semi-Supervision.} Semi-supervision reduces the label need by leveraging unlabeled data~\cite{yang2022survey}. Researchers train deep learners on a small subset of labeled data, and then produce pseudo labels on unlabeled data~\cite{sohn2020fixmatch}. The pseudo labels are used for further self-training~\cite{sahito2022better,zoph2020rethinking}. 



In incremental learning, this corresponds to partitioning the incremental learning dataset into labeled and unlabeled splits as: $\mathcal{D}_{t} = \mathcal{L}_{t} \cap \mathcal{U}_{t}$, where $\mathcal{L}_{t}$ is the standard labeled data whereas $\mathcal{U}_{t}  = \{(x_{i,t})_{i=1}^{m_{t}} \}$ is the unlabeled subset with $m_{t}$ instances. 



In this survey, we categorize semi-supervision based incremental learners by the type of unlabeled data they use, whether from within the same dataset, an auxiliary dataset or simply test data, see Table~\ref{tab:summary}. 




\vspace{-1.5mm}

\partitle{Few-shot-Supervision.} Few-shot learning reduces the label need to only a few-exemplars per-category, such as 1-shot or 5-shots. Prominent works either resort to meta-learning~\cite{vinyals2016matching,snell2017prototypical} or build upon a deep pre-trained feature extractor~\cite{tian2020rethinking}. 


In Few-Shot Class Incremental Learning (FSCIL), the objective is to update a pre-trained classifier with incrementally arriving classes with only few-exemplars, while maintaining performance on the pre-trained classes~\cite{mazumder2021few,tao2020few}. Formally, researchers introduce an initial (non-incremental) pre-training task $\mathcal{D}_{0}=\{(x_{i,0}, y_{i,0})_{i=1}^{n_{0}} \}$ with abundant data-label pairs ($n_{0} >> 5$) whereas the subsequent task only has few instances (\ie 5-shots, such as $n_{t>0} = 5$). 

In this survey, we categorize few-shot-supervised incremental learners by their core method, as in graph-based, clustering-based or architecture-based, see Table~\ref{tab:summary}. 


\vspace{-1.5mm}

\partitle{Self-Supervision.} Self-supervision omits the need for label-supervision by designing pre-text tasks~\cite{mocov2}. A promising direction is contrastive learning, where the deep learner has to pull the features of the original input and its augmented version (view) closer while pushing all the other features away. Prominent examples include SWAV~\cite{swav}, MOCO~\cite{mocov2}, and BarlowTwins~\cite{barlowtwins}. 

In incremental learning, self-supervision corresponds to training only the feature extractor $f_{\Theta}(\cdot)$ solely on unlabeled data $\mathcal{U}_{t}  = \{(x_{i,t})_{i=1}^{m_{t}} \}$. 



In this survey, we categorize self-supervised learners by \textit{how} it is being leveraged in incremental learning, whether for pre-training, auxiliary training or as the main (sole) training objective, see Table~\ref{tab:summary}. 




%% file: table1.tex
\begin{table*}[t]
\begin{center}
\begin{tabular}{llll}

\toprule

 & Setting &  Supervision &   Reference \\  
\midrule 

 
 
 Incremental Learning (IL) &   &  Label-only   &  LwF~\cite{li2017learning}\\  

\cdashlinelr{1-4}

\rowcolor{lightcyan}
Semi-Supervised IL   & Within-data &  Pseudo \& Label &  CNLL~\cite{baucum2017semi} \\   
\rowcolor{lightcyan}
Semi-Supervised IL & Auxiliary-data &   Pseudo \& Label  &  DMC~\cite{zhang2020class}\\   
\rowcolor{lightcyan}
Semi-Supervised IL & Test-data  &  Pseudo-only  & CoTTA~\cite{wang2022continual} \\   
\rowcolor{etonblue}
Few-shot-Supervised IL & Graph-based  &  Label-only (Few)  &  TOPIC~\cite{tao2020few}\\ 
\rowcolor{etonblue}
Few-shot-Supervised IL & Clustering-based  &  Label-only (Few)  &IDL-VQ~\cite{chen2020incremental}\\ 
\rowcolor{etonblue}
Few-shot-Supervised IL & Architectural-based  &  Label-only (Few)  & FSLL~\cite{mazumder2021few}\\ 
\rowcolor{blond}
Self-Supervised IL & Pre-training &  Label-only &  SSL-OCL~\cite{gallardo2021self} \\    
\rowcolor{blond}
Self-Supervised IL & Auxiliary-training & Self \& Label &  PASS~\cite{zhu2021prototype}\\  
\rowcolor{blond}
Self-Supervised IL & Main-training  &  Self-only & CaSSLe~\cite{fini2022self}\\    

\bottomrule
\end{tabular} 
\end{center}
   \caption{Towards reducing manual human supervision of incremental learners via semi-, few-shot- and self-supervision. We list three subgroups we identify for each method, in terms of data, method or training setting. For each group, we provide the type of supervision(s) necessary to train the particular incremental learner.} 
\label{tab:summary}
\end{table*}


%% file: 3-semi.tex
\section{Semi-Supervision for Incremental Learning}
\label{sec:semi}

Incremental learners leverage semi-supervision with different forms of pseudo-supervision, see Table~\ref{tab:semisl}. Pseudo-supervision serves the purpose of memory replay to reduce forgetfulness of the previous categories. 

Researchers replay labels-only~\cite{baucum2017semi,smith2021memory}, data-only~\cite{gongnote}, label and data~\cite{wang2021ordisco,brahma2021hypernetworks} and finally gradients~\cite{luo2022learning}. 

Here, we group semi-supervision-based incremental learners by their definition of unlabeled data: A subset of the target training set (Within Data), an auxiliary dataset such as from the Web, or simply test data. Within-data learners start from scratch, whereas others build upon a pre-trained neural network.

\begin{table}[t]
\begin{center}
\begin{adjustbox}{max width=0.5\textwidth}
\begin{tabular}{lccc}

\toprule 

Algorithm & Data & Pre-training & Replayed Entity \\ 
\midrule
CNNL & Within & \xmark & Pseudo-labels \\ 
DistillMatch & Within & \xmark & Pseudo-labels \\ 
ORDisCo & Within & \xmark & Pseudo-labels \& Data \\ 
MetaCon & Within & \xmark & Pseudo-labels \& Data \\ 
PGL & Within & \xmark & Pseudo-gradients \\ 
DMC & Auxiliary & \checkmark & Pseudo-labels \\ 
CIL-QUD & Auxiliary & \checkmark & Pseudo-labels \\ 
CoTTA & Test & \checkmark & Pseudo-labels \\ 
NOTE & Test & \checkmark & Data \\

\bottomrule
\end{tabular}
\end{adjustbox}
\end{center}
   \caption{Incremental Learning with Semi-Supervision.}
\label{tab:semisl}
\end{table}
\


\subsection{Learning from Within Data}




\partitle{\textbf{CNNL.}} Continuous neural network learning~\cite{baucum2017semi} is one of the early works at the intersection of semi-supervision and incremental learning. The authors train a vanilla CNN on the labeled set, which is then used to generate pseudo-labels on the unlabeled dataset. Finally, they fine-tune their incremental learner on the pseudo-labels for self-training. 



\partitle{\textbf{DistillMatch.}} DistillMatch~\cite{smith2021memory} follows a knowledge-distillation procedure, where the predictions over the unlabeled data are distilled between the current and the previous model. The authors further optimize an out-of-distribution detector to identify data points sufficiently different from the current incremental learning task (\ie past examples). This way, they mitigate forgetfulness of the previous categories.   



\partitle{\textbf{ORDisCo.}} Online replay with discriminator consistency~\cite{wang2021ordisco} follows a generative replay strategy to replay both the data and the labels. The authors leverage labeled data to train a conditional GAN generator, and leverage unlabeled data as additional examples for real-fake discrimination. To improve consistency across incremental learning tasks, the authors penalize abrupt changes within discriminator weights.



\partitle{\textbf{MetaCon.}} Meta-Consolidation~\cite{brahma2021hypernetworks} extends the generative replay scheme of ORDisCO to meta-learning setting. Instead of directly training a conditional GAN for generative replay, the authors instead optimize a conditional hyper-network~\cite{ha2016hypernetworks} that generates GAN weights. The authors use the semantic word embedding of the current task as the condition, and parameterize the hyper-network as a Variational Auto-Encoder~\cite{kingma2013auto}.  To stabilize the hyper-network, they store the first-order statistics of the incremental learning classes to replay. 





\partitle{\textbf{PGL.}} Pseudo Gradient Learners~\cite{luo2022learning} moves away from the pseudo-labels, and instead (meta-)learns to predict gradients per-input. The authors claim that the use of pseudo-labels puts too much pressure on the classifier, leading to error accumulation and performance degradation over time. By predicting gradients instead of labels, the model is not tied to a pre-defined set of classes as in pseudo-labelling, and is able to leverage out-of-distribution data to improve performance. 



\subsection{Learning from Auxiliary Data}

\partitle{\textbf{DMC.}} Deep Model Consolidation~\cite{zhang2020class} is one of the earliest to leverage unlabeled auxiliary data to mitigate forgetfulness. The authors first train a deep classifier on the labeled set, which is then used to generate pseudo-labels over auxiliary data. Pseudo-labels act as a regularizer between the current and the previous model to reduce forgetfulness of previously seen classes.  



\partitle{\textbf{CIL-QUD}.} Class-Incremental Learning with Queried Unlabeled Data~\cite{chen2022queried} builds upon DMC, however, instead takes on a retrieval-based approach. The authors store few-instances per-incremental class within the memory as query anchors. The query anchors are used to retrieve visually similar exemplars within the auxiliary dataset, which are then used for memory replay. 



\subsection{Learning from Test Data}


\partitle{\textbf{CoTTA.}} Continual Test-Time Adaptation~\cite{wang2022continual} proposes to adapt a pre-trained deep classifier at inference time. Such adaptation may be needed especially when the test data diverges from the original training source, and the source data is no longer available (imagine a self-driving car driving through changing weather conditions within a city). To tackle this novel problem, the authors propose a regularization-based approach, where they enforce consistency regularization across multiple augmentations of the same input, as well as selective fine-tuning of a few network parameters. 







\partitle{\textbf{NOTE.}} NOn-i.i.d. TEst-time adaptation~\cite{gongnote} extends COTTA to realistic scenes, where the subsequent test examples carry high temporal correlation, such as in the case of self-driving cars. The authors claim in such cases, relying on batch normalization statistics in the form of pseudo-labels may severely bias the learner towards the current batch. 
To that end, they utilize instance normalization instead of batch normalization, leading to far greater performance in comparison to CoTTA-like baselines. 



%% file: 4-few.tex
\section{Few-shot-Supervision for Incremental Learning}
\label{sec:few}

\begin{table}[t]
\begin{center}
\begin{adjustbox}{max width=0.5\textwidth}
\begin{tabular}{lcccc}

\toprule 

Algorithm & Method & Regularization & Replay & Semantic \\  
\midrule 

\textsc{TOPIC}  & Graph & Anchor Loss & \xmark & \xmark \\ 
\textsc{CEC} & Graph & \xmark & \xmark & \xmark \\ 

\textsc{IDL-VQ} & Clustering & Center Loss & \checkmark & \xmark \\ 
\textsc{SA-KD} & Clustering & \xmark & \checkmark & \checkmark \\ 
\textsc{Sub-Reg} & Clustering & $\ell_{1}$ Loss & \checkmark & \checkmark \\ 
\textsc{FACT} & Clustering & Augmentation & \xmark & \xmark \\ 

\textsc{FSLL} & Architectural & $\ell_{1}$ Loss & \xmark & \xmark \\ 
\textsc{C-FSCIL} & Architectural & Orthogonal Loss & \checkmark & \xmark \\

\bottomrule
\end{tabular}
\end{adjustbox}
\end{center}
   \caption{Incremental Learning with Few-shot-Supervision.}
\label{tab:fscil}
\end{table}






We summarize incremental learners with few-shot-supervision in Table~\ref{tab:fscil}. Few-shot-supervised learners update a pre-trained base classifier with few-examples from novel categories during incremental training. In this regard, few-shot-supervised methods tackle two fundamental challenges simultaneously: 

\textit{i) Overfitting to Novel Categories:} Optimizing for the novel classifier weights from scratch may over-fit on the few training exemplars. To that end, the authors propose to associate already learned base classifier weights with relevant novel classes during incremental training. We group the learners according to their machinery to learn base-novel class association, as in graph~\cite{tao2020few,zhang2021few}, clustering~\cite{chen2020incremental,cheraghian2021semantic,akyurek2021subspace,zhou2022forward} or the architecture~\cite{mazumder2021few,hersche2022constrained}. Some techniques also leverage semantic word embeddings to identify semantically relevant categories for transfer~\cite{akyurek2021subspace,cheraghian2021semantic}.


\textit{ii) Forgetfulness of Base Categories:} Learning novel categories may drift base classifier weights, eventually degrading the accuracy of these classes. Two effective techniques to mitigate performance degradation includes regularization via metric learning objectives such as anchor loss~\cite{tao2020few}, or simply memory replay of base class data. 






\subsection{Graph-Based Methods}






\partitle{\textbf{TOPIC}.} TOpology-Preserving knowledge InCrementer~\cite{tao2020few} is a graph-based incremental few-shot learner. The authors treat each incremental class as a novel node to be inserted to an already existing fully connected graph of base classifier embeddings. They propagate information from base classifier nodes towards novel classifier nodes with respect to their pairwise relation, as measured by the similarity of classifier embeddings. To prevent the graph from drifting and to preserve the initial graph topology, they leverage metric learning in the form of anchor loss for regularization.

\partitle{\textbf{CEC}.} Continually Evolving Classifier~\cite{zhang2021few} builds upon TOPIC, and instead leverages Graph Attention Networks~\cite{velivckovic2017graph}. In this regard, the initial graph continually evolves with the incoming stream of few-shot learning tasks. A novel incremental learning task \textit{attends} on already existing, semantically relevant categories to build the classifier weights. The authors further generate pseudo-incremental learning tasks from the base category data, which enhances novel class learning ability. The emphasis of CEC is more on the novel class learning rather than maintaining base class performance, which eventually exacerbates forgetfulness.   





\subsection{Clustering-Based Methods}

\partitle{\textbf{IDL-VQ}.} Incremental Deep Learning Vector Quantization~\cite{chen2020incremental} utilizes Gaussian-Mixtures to quantize visual features of learned categories to reference vector centroids. Then, any incoming novel class is represented by their soft similarity with existing reference vectors. The authors further store 1-shot per-class for replay to reduce forgetfulness.  






\partitle{\textbf{SA-KD}.} Semantic-Aware Knowledge Distillation~\cite{cheraghian2021semantic} uses K-means clustering instead to build reference class centroids, which are used to represent a novel input. The representation is then projected into semantic word embedding space to further promote base-novel class association.





\partitle{\textbf{SUB-REG}.} Subspace-Regularization~\cite{akyurek2021subspace} utilizes QR-decomposition to project base classifier embeddings to an orthogonal sub-space. They then measure soft similarities between a novel class input and sub-space vectors to represent the novel class weights. The authors additionally apply $\ell_{1}$ loss to penalize abrupt weight changes within the classifier. 






\partitle{\textbf{FACT}.} Forward-Compatible Training~\cite{zhou2022forward} is the current state-of-the-art in few-shot incremental learning. Authors first showcase that the feature space of few-shot learners is fully occupied  by pre-training (base) classes, leaving no room for future (few-shot) classes, hence limiting forward-compatibility. To that end, they propose to simultaneously assign an input image to a separate cluster orthogonal to the base classes, effectively reserving room for novel categories. Combined with mixup augmentations, FACT improves performance on both base and novel classes.  






\subsection{Architectural Methods}


\partitle{\textbf{FSLL}.} Few-shot Lifelong Learning~\cite{mazumder2021few} selects a few weights from the architecture to fine-tune per-incremental learning task. In doing so, the authors simultaneously prevent overfitting by limiting the model capacity, while preventing forgetting via minimizing interference across learning tasks. Combined with $\ell_{1}$ loss, FSLL significantly improves over TOPIC across several benchmarks. 



\partitle{\textbf{C-FSCIL}.} Constrained-FSCIL~\cite{hersche2022constrained} is a recent technique that operates on classifier embeddings (prototypes) instead. The authors expand the architecture with novel classifier embeddings with incoming stream of tasks. In doing so, they impose quasi-orthogonality across learned embeddings, effectively minimizing task interference, hence less forgetful. 

%% file: 5-self.tex
\section{Self-Supervision for Incremental Learning}
\label{sec:self}

We summarize incremental learners that leverage self-supervision in Table~\ref{tab:ssl}. We identify three different trends in self-supervision for incremental learning:  
\textit{i) Pre-training:} To pre-train the backbone prior to incremental learning, to warm-start with discriminative weights~\cite{gallardo2021self}. This line of research builds upon the idea that self-supervised pre-training reduces the need for label-supervision in subsequent (transfer) learning tasks.  
\textit{ii) Auxiliary-training:} To supplement standard label-supervision with self-supervised objectives during training to obtain a more discriminative feature space~\cite{zhu2021prototype}. This line of research builds upon the idea that self-supervised learning tasks can provide additional supervisory signals to the learner to prevent overfitting. 
\textit{iii) Main-training:} To train solely based on self-supervised learning objectives~\cite{purushwalkam2022challenges,madaan2021representational,fini2022self,gomez2022continually}, which are then evaluated by linear probing after training. These learners build upon the idea that self-supervision can replace label-supervision to induce a discriminative deep feature extractor.

\begin{table}[t]
\begin{center}
\begin{adjustbox}{max width=0.5\textwidth}
\begin{tabular}{lccc}

\toprule 


Algorithm  & Setting & Self-Supervision \\ 
\midrule 
SSL-OCL & Pre-training & MOCO/SwAV \\ 
PASS & Auxiliary-training & SLA \\  
Buffer-SSL & Main-training & SimSiam \\ 
LUMP   & Main-training & SimSiam/Barlow-Twins \\  
CaSSLe & Main-training & SimCLR/Barlow-Twins/etc. \\  
PFR          & Main-training & Barlow-Twins  \\

\bottomrule
\end{tabular}
\end{adjustbox}
\end{center}
   \caption{Incremental Learning with Self-Supervision.}
\label{tab:ssl}
\end{table}
\

\vspace{-8mm}
\partitle{Pre-training.} Self-Supervised Learning for Online Continual Learning (\textbf{\textsc{SSL-OCL}})~\cite{gallardo2021self} proposes to pre-train the backbone weights prior to incremental training via self-supervision. This way, the authors aim to leverage transfer-learning abilities brought by pre-training dataset. Specifically, they evaluate MoCo-v2~\cite{mocov2}, Barlow-Twins~\cite{barlowtwins} and SwAV~\cite{swav} for self-supervised pre-training. The authors conclude that in contrast to standard label-supervised pre-training, self-supervised pre-training is always superior, and SwAV consistently outperforms the compared alternatives.

\partitle{Auxiliary-training.} Prototype-Augmented Self-Supervision (\textbf{\textsc{PASS}})~\cite{zhu2021prototype} is a regularization-based incremental learning technique. The model optimizes a single-prototype per-incremental class, where the prototype is learned by standard label-supervision. To improve generalization and avoid overfitting, the authors resort to augmentation. Specifically, they use Self-Supervised Label Augmentation (SLA)~\cite{lee2020self} to generate four-fold rotations of the original input ($[0, 90, 180, 270]$), which are then used as additional pre-text tasks to differentiate for the model. The authors show such method mitigates overfitting to the previously learned classes, leading to superior results. 



\subsection{Main-training}
\label{sec:self-main}

Using self-supervision as the sole supervision signal for incremental training is probably the most promising direction, as it requires no labels at training times. For this, we identify two memory-based and two regularization-based approaches. 






\partitle{\textbf{Buffer-SSL}.} Buffer Self-Supervised Learning~\cite{purushwalkam2022challenges} is a memory-based approach that extends an off-the-shelf self-supervision algorithm, SimSiam~\cite{simsiam} for the case of incremental learning. In doing so, the authors identify three main challenges. First, SimSiam has no mechanism to retain information for previously seen data, leading to severe forgetting. To mitigate this, the authors supplement SimSiam with a memory buffer for replay. A trivial approach would be to store all examples within the memory, however, results in a large memory size, and there is a high redundancy across subsequent learning frames (such as in a video). To that end, the authors only store cluster centroids, which reduces memory size and increase memory variability. Using such buffer is shown to significantly reduce forgetfulness of self-supervised representations.


\partitle{\textbf{LUMP.}} Lifelong Unsupervised Mixup~\cite{madaan2021representational} is a memory-based approach utilizes mixup augmentation to mitigate forgetfulness. Specifically, the authors learn to mixup the input instances with those from the past learning tasks stored within the memory. Replaying such examples effectively reduces forgetting and improves performance. The authors demonstrate their algorithm with both SimSiam~\cite{simsiam} and Barlow-Twins~\cite{barlowtwins}, where both algorithms lead to similar performance.



\partitle{\textbf{CaSSLe}.} CaSSLe is a regularization-based approach to self-supervised incremental learning~\cite{fini2022self}. Since storing data from past learning tasks is memory inefficient and may violate privacy, the authors instead learn to distill self-supervised representations between the current and the past model. The distillation is performed in a predictive manner, where the current model's features are projected onto the previous model's feature space. The authors apply their method on several different self-supervised learning algorithms, observing similar performance. The authors propose to go beyond class-incremental setting, and also evaluate data-incremental (the data is partitioned randomly regardless of the classes) and domain-incremental (data is partitioned by domain label, such as \ie real $\rightarrow$ sketch $\rightarrow$ clipart). Regardless of the setting, self-supervised representations are found to be always more accurate and less forgetful than label-supervised counterparts, which is promising for reducing the need for label-supervision in incremental learning.  

\partitle{\textbf{PFR}.} Projected Functional Regularization~\cite{gomez2022continually} is a regularization-based technique, very similar to CaSSLe. The authors extend Barlow-Twins with a distillation-based objective. Specifically, they learn to project the current visual representation to the previous model representation. The authors showcase that Barlow-Twins with PFR objective exhibits lower forgetting and higher accuracy. 

%% file: 6-conclusion.tex
\section{Conclusion and Future Directions}
\label{sec:conclusion}



In this survey, we establish the lack of label-efficiency as a major bottleneck in deploying realistic incremental learners. We unify three different ways to improve label-efficiency, namely semi, few-shot and self-supervised learning. 

Although promising, these set of learners are not without limitations. To that end, to inspire future research, in this section we first identify limiting factors for label-efficiency and methodology. We then conclude with novel directions to explore for label-efficient incremental learning. 


\subsection{Limitations and Alternative Methods}




\partitle{Semi-Supervision.} Incremental learning via semi-supervision leverages partially labeled data in the form of pseudo-supervision. The quality of pseudo-supervision is partly determined by the amount of labeled data. To that end, the models still require a significant amount of labels to be provided to work well, limiting their label-efficiency. 

Also, the level of pseudo-labelling noise may accumulate over time, especially for long incremental learning sequences, limiting the model performance~\cite{wang2022continual}. A potential remedy is to move from pseudo-labels to pseudo-gradients~\cite{luo2022learning}, however the gradient estimation may also be suboptimal by time. To that end, we see promise in updating normalization parameters via input statistics \textit{without} any form of pseudo-supervision~\cite{gongnote}.



\partitle{Few-shot-Supervision.} Few-shot-supervised learners make use of labels both pre- and during incremental training. The label complexity of pre-training is especially huge, since it requires many shots and many categories relevant to novel few-shot classes. In data hungry fields like medical imaging or visual anomaly detection, limiting their applicability. 

Additionally, few-shot-supervised learners freeze the backbone during incremental learning to prevent over-fitting. While working well, such practice is unnatural, since humans can leverage few-shots for learning. One potential remedy would be to rely on Sharpness-Aware Minimization, as few-shot learning is shown to exhibit loss landscape with sharp and poor local minimum~\cite{abbas2022sharp}. 



\partitle{Self-Supervision.} The label complexity of self-supervised incremental learners are on par with vanilla incremental learners, when self-supervision is used for pre-training or for auxiliary-training. Using self-supervision as the sole supervisory signal holds the key to completely omit the labels during training. However, purely self-supervised incremental learning requires a separate labeled linear-probing stage for evaluation purposes, limiting their use cases. 

Also, self-supervision for incremental learning solely relied on contrastive learning-based approaches to extract supervisory signals~\cite{mocov2,barlowtwins}. However, recent studies like MAE show promise of reconstruction-based objectives over contrastive-based counterparts, which we believe is worthy of exploration~\cite{he2022masked}.

\subsection{Novel Problems} 

Inspired by our survey, here we recommend novel problems to investigate in future research to build more realistic incremental learners. 



\partitle{Mixed-Supervised Learning.} In this survey, we show that researchers follow three main directions to reduce the need for supervision in a \textit{disjoint} manner. However, a combination of different forms of supervision is common in non-incremental learning, such as the combination of self- and few-shot-supervision~\cite{su2020does}, or semi and few-shot-supervision~\cite{ren2018meta,li2019learning}. 

To that end, our first recommendation is to explore such mixed supervision settings for incremental learning. 

\partitle{Incremental Dense Learning.} We limit our survey to the fundamental task of image classification. However, for incremental dense prediction tasks such as image segmentation~\cite{maracani2021recall}, the demand for labels is explosive. For image segmentation, annotators label each and every pixel in a high-resolution scene over thousands of imagery. 



We believe future research should incorporate different ways to reduce the need for incremental supervision for dense prediction tasks, as the demand for building commercial incremental learners grows. We note that the initial effort(s) have been made for weakly-supervised incremental object segmentation, however with limited performance over non-incremental counterparts~\cite{cermelli2022incremental}. 



\partitle{Incremental Active Learning.} In active learning, the learning agent selects the most influential examples to be annotated by a human expert~\cite{munjal2022towards}. This significantly reduces the cost of labeling, as only a small portion of the examples suffice to train a good model. 

Surprisingly, active learning has found little to no application in incremental learning agents. However, such method can help to select a small set of exemplars to annotate for semi- or few-shot-supervised incremental learning, either for pre-training or incremental-training.


\partitle{Incremental Object Discovery.} Humans exhibit an astounding ability to discover never-before-seen objects, with little to no supervision. We are able to group instances of novel objects with little effort. However, existing incremental learners currently has no novel object discovery capacity~\cite{han2020automatically}, as all the objects are at least partially labeled. This is unrealistic, since the visual world always presents novel objects in daily life, partially thanks to the advances in technology.


To that end, we believe that label-efficient learners should be capable of not only learning from limited supervision, but also discover novel objects. 




To conclude, we believe label-efficiency is a key factor to build autonomous, human-like life-long learning agents. The potential in leveraging the ever-growing, massive-scale unlabeled data is yet to come, and we hope our survey provides a good introduction to this important field of newly emerging research. 














\section{Acknowledgements}

Mert Kilickaya's research is fully funded by ASM Pacific Technology (ASMPT).